# SEMANTIC DESCRIPTIONS OF 24 EVALUATIONAL ADJECTIVES, FOR APPLICATION IN SENTIMENT ANALYSIS

*Cliff Goddard, Maite Taboada, Radoslava Trnavac*

Griffith University, Simon Fraser University

c.goddard@griffith.edu.au, mtaboada@sfu.ca, rtrnavac@sfu.ca

**Abstract**: We apply the Natural Semantic Metalanguage (NSM) approach (Goddard & Wierzbicka 2014) to the lexical-semantic analysis of English evaluational adjectives and compare the results with the picture developed in the Appraisal Framework (Martin & White 2005). The analysis is corpus-assisted, with examples mainly drawn from film and book reviews, and supported by collocational and statistical information from WordBanks Online. We propose NSM explications for 24 evaluational adjectives, arguing that they fall into five groups, each of which corresponds to a distinct semantic template. The groups can be sketched as follows: "First-person thought-plus-affect", e.g. *wonderful*; "Experiential", e.g. *entertaining*; "Experiential with bodily reaction", e.g. *gripping*; "Lasting impact", e.g. *memorable*; "Cognitive evaluation", e.g. *complex, excellent*. These groupings and semantic templates are compared with the classifications in the Appraisal Framework's system of Appreciation. In addition, we are particularly interested in sentiment analysis, the automatic identification of evaluation and subjectivity in text. We discuss the relevance of the two frameworks for sentiment analysis and other language technology applications.

This report is a supplementary document to Goddard, Taboada and Trnavac (2016), henceforth GTT. It contains NSM semantic explications for attributive uses of 24 evaluational adjectives, accompanied by brief comments or supporting observations. The assumptions and operating principles of the NSM approach (cf. Goddard (2011), Goddard and Wierzbicka (2014)) are explained in GTT, along with the analytical procedure that led to the explications proposed below. The analysis drew on corpus data from WordsBanks Online, a commercially available corpus service. For background to the study of evaluational language generally, and its application to sentiment analysis, see Martin and White (2005), and Taboada et al. (2011).

Evaluational adjectives, and the language of evaluation generally, pose fascinating challenges for semantic description, both on account of their inherent subjectivity and because of the sheer number of subtly different meanings involved. For the same reasons, they pose special challenges for computational linguistics and affective computing, including for sentiment analysis (Hudlicka 2003; Taboada et. al 2011; Trnavac & Taboada 2012).

The primary goals of the GTT paper are two-fold. The first goal is to apply the Natural Semantic Metalanguage (NSM) approach to a selection of evaluational adjectives. The NSM approach is a well developed approach to lexical-semantic analysis, based on reductive paraphrase (Wierzbicka 1996; Goddard & Wierzbicka 2014; Peeters 2006; Goddard 2011; Levisen 2012; and other works). There is a large "back catalogue" of NSM studies into the evaluative lexicon, especially in the domains of emotion and values (e.g. Wierzbicka 1999; Harkins & Wierzbicka 2001), but this is the first NSM study of evaluational adjectives. We present and discuss original NSM explications for a total of 39 such adjectives (15 in the GTT paper; 24 in this report), arguing that they fall into five sub-groups, each conforming to a distinct structure or semantic template. This selection has not been chosen at random, but is a subset of about 40-50 adjectives currently under study.

Our second goal is to compare these results with the picture developed within the Appraisal Framework (Martin & White 2005; Martin in press; among others). This is arguably the most influential approach to evaluational language, having been developed over the last 15 years under the auspices of Systemic Functional Linguistics (SFL) (Halliday 1985; Halliday & Matthiessen 2014). SFL follows the structuralist tradition insofar as it is based on a system of classifications and oppositions.

The five groupings that have emerged from the process of NSM analysis are listed in Table 1, with sample adjectives for each grouping. In the Appraisal Framework (Martin & White 2005), they fall into the category of Appreciation, which has a number of subtypes as discussed later. In Table 1, each grouping has two rows, one for positive and one for negative adjectives.

**Table 1: Five groupings of evaluational adjectives († = discussed in the GTT paper)**

A+: †*great*, †*wonderful*, †*terrific*, *fabulous*, *awesome*
A–: *terrible*, *awful*, *dreadful*

B1+: †*entertaining*, †*delightful*, *fascinating*, *compelling*, *interesting*, *touching*
B1–: *boring*, *predictable*

B2+: †*gripping*, †*exciting*, *tense*, *suspenseful*, *stunning*
B2–: *disgusting*, *sickening*

C+: †*powerful*, †*memorable*, *haunting*, *inspiring*
C–: *disturbing*, *depressing*

D+: †*complex;* †*excellent,* †*outstanding;* †*impressive;* †*brilliant, original, clever*
D–: *disappointing; dismal, woeful*

——————————————————————————————————————

In the rest of this report, we include explications for 24 of the adjectives in Table 1. The templates are listed in the same order as in GTT.

# 1 TEMPLATE A WORDS

Words falling under Template A, e.g. *great, wonderful, terrific, awesome, fabulous, terrible, awful, dreadful*, can be characterised as "first-person thought-plus-feeling" words. These words are overtly subjective, modelled in the explications by the lead component 'I think about it like this: …'. Then follows a model thought, which in this set of explications begins with a strong evaluation: either 'this X is very good' or 'this X is very bad'. The special character of each evaluation comes from the thought component, which is different for each adjective. The template is completed with a component indicating that on account of thinking as he/she does, the speaker feels 'something very good' or 'something very bad', as the case may be.

 *Great, wonderful,* and *terrific* are explicated in GTT. Explications for five additional Template A words follow.

## 1.1 Template A+

*an awesome X,* e.g. *an awesome movie*

| I think about it like this: |
|---|
|  "this X is very good |
|  people can think like this: 'it can't be like this' " |
| when I think like this, I feel something very good because of it |

*a fabulous X,* e.g. *a fabulous film, holiday*

| I think about it like this: |
|---|
| "this X is very good<br>  I didn't know before that X's (= such things) can be like this<br>  I want to say more about it, at the same time I don't know what I can say" |
| when I think like this, I feel something very good because of it |

Impressionistically, *fabulous* feels close to *wonderful*, but there are collocations where one or the other is strongly preferred or which convey different implications; compare *a wonderful opportunity* and *?a fabulous opportunity*. *Fabulous* sounds enthusiastic, breathless, deliberately hyperbolic.

## 1.2   Template A–

*a terrible X,* e.g. *a terrible movie; a terrible mistake*

| I think about it like this: |
|---|
| "this X is very very bad<br>    something very bad can happen because of this" |
| when I think like this, I feel something very bad because of it |

*an awful X,* e.g. *an awful movie, an awful job, an awful stay*

| I think about it like this: |
|---|
| "this X is very bad<br>    something very bad can happen to someone because of this" |
| when I think like this, I feel something very bad because of it |

*a dreadful X,* e.g. *a dreadful mistake, a dreadful lie, a dreadful outcome*

| I think about it like this: |
|---|
| "this X is very bad<br>    something very bad can happen to people because of this" |
| when I think like this, I feel something very bad because of it |

• *Awful* seems more subjective, more personalised than *dreadful*; cf. *That's awful for you* vs. *\*That's dreadful for you.* Similarly, a sentence like *I've got an awful pimple* seems quite ordinary, but would be odd with *dreadful*. • There are collocational indications that the thought behind *dreadful* is broader and less personal than for *awful*, e.g. descriptions of the climate tend to sound better with *dreadful* than with *awful*.

## 2 TEMPLATE B WORDS

Words falling under the B Templates (subtypes B1 and B2) are termed "experiential" evaluators. Examples include: *entertaining, delightful* for B1, *gripping, exciting* for B2. They differ from the Template A words in several ways. First, they are less overtly subjective. This is modelled in the explications by a component saying that 'someone can think like this (about it): ...'. That is, these evaluational words in this group work by invoking a hypothetical 'someone' and attributing certain thoughts and associated feelings to this hypothetical someone. Second, words in this group say something about someone's "experience" of the things being evaluated. Briefly, this means that the thoughts and feelings being depicted are associated with the time period during which someone experiences (watches, reads, or is otherwise exposed to) the stimulus. Third, the B2 Template includes an additional semantic component alluding to a potential bodily effect on the experiencer.

The notation => indicates that the details of the top-most section of the explications are not spelt out in full (mainly because they vary somewhat depending on the nature of the noun). See GTT for more detail.

*Entertaining, delightful, gripping* and *exciting* are explicated in GTT. Explications for ten additional Template B1/B2 words follow.

### 2.1 Template B1+

*an interesting –,* e.g. *an interesting documentary, an interesting read,* =>

| |
|---|
| during this time (e.g. when this someone watches this film, reads this book; when certain things happen to this someone), |
|    this someone can think like this at many times: |
|     "I want to know more about this |
|      it can be good if someone says some things about it |
|      it can be good if I think about it for some time" |
| when this someone thinks like this, he/she can feel something because of it |
|    not like people feel at many other times |

• The word *interesting* implies not only wanting to know more, but also aspects of discursive engagement and cognitive engagement (cf. Goddard in press).

a *compelling$_1$ –,* e.g. *a compelling performance, a compelling story* =>

| |
|---|
| during this time (e.g. when this someone watches this film, reads this book; when certain |
| things happen to this someone), |
| this someone can think like this at many times: |
| "something is happening now<br>  I want to know what will happen after this, I can't not (= have to) know it" |
| when this someone thinks like this, he/she can feel something because of it<br>  not like people feel at many other times |

The word *compelling* has two distinct meanings: the "experiential" meaning explicated above, and another purely cognitive meaning, as in phrases like *compelling evidence, a compelling reason, argument,* etc. Only the "experiential" meaning (which, incidentally is highly language-specific) can occur in the frame *It's compelling.*

*a fascinating –,* e.g. *a fascinating exploration, conversation* =>

|  |
|---|
| during this time (e.g. when this someone watches this film, reads this book; when certain things happen to this someone), |
| this someone can think like this at many times: |
| "I want to know more about this, I want it very much |
| when this someone thinks like this, he/she can feel something very good because of it |

There is an intuition that if something is *fascinating*, we are finding out something new and we want more (i.e. *fascinating* implies *very interesting*, at least from a cognitive point of view), but also that *fascinating* is somehow pleasurable, as captured in the final component.

a *touching –,* e.g. *a touching story, memoir, ballad* =>

|  |
|---|
| during this time (e.g. when this someone watches this film, reads this book; when certain things happen to this someone), |
| this someone can think like this at many times: |
| "someone did something a moment ago, not like people do at many times<br>  because of this, I know that this someone feels something very good towards<br>    someone else at this time" |
| when this someone thinks like this, he/she can't not feel something good for a short time |

• The word *touching* is not inherently durational (cf. expressions like *a touching moment, a touching gesture*), but we explicate it above in a durational frame. • Describing something as *touching* implies a more or less immediate reaction to an act that shows someone's strong affection towards someone else. It is akin to *heart-warming*. • On a point of detail, the final 'feeling' component contains a time adjunct that states that the feeling is short-lived.

## 2.2 Template B1–

*a boring —,* e.g. *a boring lecture, boring meetings* =>

| during this time (e.g. when this someone watches this film, reads this book; when certain things happen to this someone), |
|---|
| this someone can think like this about it: |
| "this is like many things were before |
| I don't want it to be like this |
| I want to do something else now" |
| when someone thinks like this, he/she can feel something bad because of it |

In some ways, *boring* is a semi-converse to *interesting*, but it is less sophisticated. Children use the word *boring* a lot earlier than *interesting*. It implies that someone is "attending" to what's happening and finds it wanting.

*a predictable —,* e.g. *a predictable storyline, ending, response*

| during this time (e.g. when this someone watches this film, reads this book; when certain things happen to this someone), |
|---|
| this someone can think like this at many times: |
| "something is happening now |
| I knew before that this would happen" |
| when someone thinks like this, he/she can feel something bad because of it |

Interestingly, this explication is not "prospective", i.e. it is not about having the impression of knowing what's coming next, but rather about recognising something about what is happening now (e.g. in the film or book). This is a simpler and more "experiential" meaning than one would expect from the verb *predict*, which is future-oriented.

## 2.3 Template B2+

As mentioned, this group of words follows a very similar structure to the B1 group, but with an extra component suggesting some kind of potential bodily reaction.

*a tense —*, e.g. *a tense, taut thriller; a tense scene*

| during this time (e.g. when this someone watches this film, reads this book; when certain things happen to this someone), |
|---|
| this someone can think like this at many times: |
| "something very bad can happen after a short time |
| I don't want it to happen |
| I can't think about other things now" |
| at the same time, this someone can think about it like this: "I know that this is not true" |
| when this someone thinks like this, he/she can feel something because of it |
|     not like people feel like at many other times |
| at the same time something can happen in this someone's body because of it |

The word *tense* is polysemous. We explicate here its specialised "experiential" meaning in contexts like *a tense movie*, *a tense scene*. In its other meanings, e.g. in expressions like *tense muscles, a tense situation*, the word implies negative feeling but when used about a movie, book, etc., the implied feeling is not negative. This is accounted for by the component capturing the experiencer's awareness of the "non-reality" of the situation. Note that whether or not a film, book, etc. is *tense* does not depend purely on unpredictability. One may know exactly what will happen but still find it *tense*.

*a suspenseful —*, e.g. *a suspenseful romance, plot*

| |
|---|
| during this time (e.g. when this someone watches this film, reads this book; when certain things happen to this someone), |
|     this someone can think like this about someone at many times: |
|     "something will happen after a very short time |
|     maybe it will be very good for this someone, maybe it will be very bad for this someone |
|     I want to know what will happen, I want to know it now" |
| when this someone thinks like this, he/she can feel something because of it |
|     not like people feel at many other times |
| at the same time something can happen in this someone's body because of it |

With *suspenseful*, the experiencer is sure that something will happen very soon, and the stakes are high, but the potential event in question does not necessarily have to be negative. *Suspense* can also come from waiting to find out whether something very good will happen, e.g. winning a competition or prize, cf. *Don't keep in me in suspense*.

*a stunning —*, e.g. *a stunning performance* =>

| |
|---|
| during this time (e.g. when someone watches this film, reads this book; when certain things happen to this someone), |
|     this someone can think like this at many times: |
|     "this is something very good, I don't know how it can be like this |
|     because of this I can't think well at this moment" |
| when someone thinks like this, he/she can't not feel something because of it |
| people don't feel like this at many times |
| at the same time he/she can feel something in the body because of it |

• Many uses of *stunning* occur in non-durational contexts, and pertain to the immediate cognitive experience of seeing something remarkable, e.g. *stunning looks,* or realising that something remarkable has happened, e.g. *a stunning victory*. Above, however, we explicate the word used in a durational context, as it pertains to the experience of someone watching a film, a performance, etc. • Data from WordBanks shows that *stunning* rarely occurs conjoined with other adjectives and almost never occurs with *very* or *extremely* (though can be modified with Focus intensifiers, e.g. *absolutely*, *quite*, *simply*).

## 2.4 Template B2–

*a disgusting –,* e.g. *disgusting behaviour, a disgusting sight*

| |
|---|
| during this time (e.g. when this someone watches this film, reads this book; when certain things happen to this someone), |
| this someone can think like this at many times: |
| "someone is doing something very bad now, something very bad is happening now because of it" |
| when someone thinks like this, he/she can't not feel something very bad because of it |
| at the same time he/she can feel something bad in the body like people feel at some times when there is something very bad inside the mouth [m] |

• This explication is simplified from Goddard (2014a). • In WordBanks, its most common attributive uses are with generic nouns like *thing* (e.g. *the most disgusting thing I've ever seen/heard*) or abstract nouns indicating human actions and behaviours, e.g. *disgusting habit/act, disgusting behaviour*. • *Disgusting* is hardly ever modified by *very*, implying that it already includes VERY in its meaning. • There are semantic links with the interjections *Yuck*! and *Ugh!* (cf. Goddard 2014).

*a sickening –,* e.g. *a sickening re-enactment; sickening cruelty*

| |
|---|
| during this time (e.g. when this someone watches this film, reads this book; when certain things happen to this someone), |
| this someone can think like this at many times: |
| "something very bad is happening to someone's body at this time I can't not think about it" |
| when someone thinks like this, he/she can't not feel something very bad because of it |
| like people can't not feel something very bad at some times when something very bad is happening inside the body |

• Like *disgusting*, the word *sickening* is hardly ever modified by *very*. • Many nouns that commonly go with *sickening*, e.g. *abuse*, *cruelty*, *attack*, and *crime*, clearly evoke human action as the cause. • Even more frequently it is found in a non-durational frame with nouns denoting sounds of someone's body undergoing a traumatic impact, e.g. *collision*, *thud*, *crack*.

## 3 TEMPLATE C WORDS

Words falling under Template C, e.g. *powerful*, *memorable*, *haunting, disturbing,* etc., are not focused on what it was like to have the experience but on the subsequent on-going effect on the viewer (reader, participant, etc.). The middle section of the explications, which model this "after effect", contains psychological components hinged around semantic primes such as THINK and FEEL. Additionally, as far as we can see, such words always imply a broad evaluation as either good or bad, which appears as the final component of the template.

*Powerful* (in its evaluational sense, e.g. *a powerful movie)* and *memorable* are explicated in GTT. Explications for four additional Template C words follow.

### 3.1 Template C+

*a haunting —*, e.g. *a haunting book/film; a haunting melody*

| when someone does something like this for some time (e.g. watches this film, reads this book, listens to this music), something happens to this someone because of it |
|---|
| because of this, for some time afterwards it is like this: this someone can't not think about it at some times when this someone thinks about it, this someone can't not feel something this someone can think about it like this: "I felt something like this some time before, I know when I felt it" when this someone thinks like this, he/she can feel something bad because of it, at the same time he/she can feel something good because of it |
| people can think about it like this: "this is good" |

*an inspiring —,* e.g. *an inspiring story, tribute, account*

| when someone does something like this for some time (e.g. watches this film, reads this book, listens to this music), something happens to this someone because of it |
|---|
| because of this, for some time afterwards it is like this: this someone can think like this: "people can do some very good things if they very much want to do these things |
| I can do some very good things if I very much want to do these things" |
| people can think about it like this: "this is good" |

## 3.2 Template C–

*a depressing ―*, e.g. *a depressing film, story, account; a depressing song*

| |
|---|
| when someone does something like this for some time (e.g. watches this film, reads this book, listens to this music),<br>   something happens to this someone because of it |
| because of this, for some time afterwards it is like this:<br>   this someone thinks like this at some times:<br>    "very bad things happen to people at many times<br>    people don't want it to be like this, at the same time they can't do anything because of it"<br>   when this someone thinks like this, he/she can't not feel something bad because of it |
| people can think about it like this: "this is bad" |

*a disturbing ―,* e.g. *a disturbing film; disturbing images*

| |
|---|
| when someone does something like this for some time (e.g. watches this film, reads this book, listens to this music),<br>   something happens to this someone because of it |
| because of this, for some time afterwards it is like this:<br>   this someone can't not think about it at some times<br>   when this someone thinks about it, he/she can't not feel something bad<br>   because of this, this someone can't think well about other things<br>    like he/she can at these times |
| people can think about it like this: "this is bad" |

## 4 TEMPLATE D WORDS

Words falling under Template D, e.g. *complex; excellent; brilliant,* are purely cognitive evaluations. That is, although they may imply feeling, they do not encode any feeling. There are several discernable sub-groups within this group, but the differences concern the nature of the semantic components involved rather than the template structure.

*Complex*, *excellent*, *outstanding*, *impressive*, and *brilliant* are explicated in GTT. Explications for five additional Template words follow.

### 4.1 Template D+

*an original –*, e.g. *an original story, a truly original idea*

| |
|---|
| if someone knows what this X is like,<br>   this someone can think about it like this: |

> "I know that there was nothing like this X before
> I know that there is something like this now because someone did something, not like
>     someone else did before
> this is good"

• The word *original* has a distinct but related meaning that we see in contexts like *The original proposal was that* ... The explication above is for the evaluative meaning (roughly, 'creative') which, it can be noted, often occurs with a modifier, e.g. *wholly, really, very*.

*a clever –,* e.g. *a clever story, a clever plan, a clever solution*

> if someone knows what this X is like,
>     this someone can think about it like this:
>     "this X is not like many X's (such things)
>     something can't be like this if someone doesn't think well about it
>         for some time before"

As reflected in the explication, calling something *clever* is not necessarily a quality endorsement.

## 4.2  Template D–

*a disappointing –,* e.g. *a disappointing film; disappointing results*

> if someone knows what this X is like,
>     this someone can think about it like this:
>     "I thought about it like this before: this can be very good
>     I felt something good because of this
>     I know now that it is not like this"

According to the explication, the word *disappointing* represents a cognitive evaluation, i.e. one that is not necessarily linked with a feeling. Relatedly, even though the word ends with *-ing* and has the appearance of a participial adjective, it implies a holistic appraisal.

*a dismal –,* e.g. *a dismal failure, a dismal situation*

> if someone knows what this X is like,
>     this someone can think about it like this:
>     "this is something very very bad
>     very few such things (= things like this) are like this
>     if people think about this, they can feel something very bad because of it"

*a woeful –,* e.g. *a woeful performance, in a woeful state*

| if someone knows what this X is like, |
|---|
|    this someone can think about it like this: |
|    "someone did something very badly before |
|    this is very bad for someone |
|    if people think about this, they can feel something very bad because of this" |

• *Dismal* and *woeful* may seem very similar but in some contexts there are clear acceptability contrasts, e.g. *a dismal/\*woeful failure; dismal/\*woeful weather*; ?*dismal/woeful tragedy.* The explications account for this by explaining *woeful* in terms of someone's very bad performance leading to a very bad "personal" consequence ('this is very bad for someone'). • In relation to the final component of both explications, it can be noted that dictionaries sometimes mention causing 'gloom', 'dismay' or 'sadness' as part of the meanings of these words.